# A systematic review on expert systems for improving energy efficiency in the manufacturing industry

Borys Ioshchikhes[a,*], Michael Frank[a], Matthias Weigold[a]

[a]Technical University of Darmstadt, Institute for Production Management, Technology and Machine Tools (PTW), Otto-Berndt-Str. 2, 64287 Darmstadt, Germany

**Abstract**

Against the backdrop of the European Union's commitment to achieve climate neutrality by 2050, efforts to improve energy efficiency are being intensified. The manufacturing industry is a key focal point of these endeavors due to its high final electrical energy demand, while simultaneously facing a growing shortage of skilled workers crucial for meeting established goals. Expert systems (ESs) offer the chance to overcome this challenge by automatically identifying potential energy efficiency improvements and thereby playing a significant role in reducing electricity consumption. This paper systematically reviews state-of-the-art approaches of ESs aimed at improving energy efficiency in industry, with a focus on manufacturing. The literature search yields 1692 results, of which 54 articles published between 1987 and 2023 are analyzed in depth. These publications are classified according to the system boundary, manufacturing type, application perspective, application purpose, ES type, and industry. Furthermore, we examine the structure, implementation, utilization, and development of ESs in this context. Through this analysis, the review reveals research gaps, pointing toward promising topics for future research.

*Keywords:* sustainability, climate neutrality, industrial processes, energy analysis, optimization

## 1. Introduction

Climate change is one of the most imperative topics of modern times. The European Climate Law addresses this threat by setting a net greenhouse gas emissions reduction target of at least -55% by 2030, compared to 1990 levels (Regulation (EU) 2021/1119, 2021). Energy use plays a critical role in the pursuit of this objective, as it was the source of almost three-quarters of global emissions in 2016. The industry sector is responsible for about 30% of emissions, with energy use accounting for 24.2%, making it the main source of emissions. (Ritchie et al., 2020) To achieve decarbonization, efforts are being made to increase the share of electricity in all sectors, and its share of final energy consumption is projected to rise from 20% today to over 50% by 2050. Since the industry sector was already the largest consumer of final electric energy in 2019, representing 41.9 %, there is an imperative urgency to take action (International Energy Agency [IEA], 2021).

In response to this challenge, companies are prioritizing energy efficiency improvements as a way to simultaneously achieve affordability, supply security, and climate goals (ABB Ltd, 2022). Energy efficiency is a measure that quantifies the utilization of energy in relation to the output or yield of services, goods, commodities, or energy (Deutsches Institut für Normung e.V., 2018). Increasing energy efficiency therefore means making better use of existing resources. According to a 2022 global survey conducted among over 2,200 industrial companies in 13 countries, 97% of the companies had either already invested or were planning to invest in energy efficiency. Moreover, 89% of these companies anticipated increasing their energy efficiency investments over the next five years, while

* Corresponding author. Tel.: +49-615-18229773; +49-6151-822-9675.
  *E-mail address:* b.ioshchikhes@ptw.tu-darmstadt.de

52% had the ambitious goal of achieving net zero within the same timeframe (ABB Ltd, 2022). However, despite the increase in energy efficiency in the manufacturing sector in recent years, there is still considerable potential for further improvement (IEA, 2022). Especially energy submetering of individual industrial processes is low, showing a deficiency of monitoring and targeting systems that are crucial for effective energy management and energy efficiency efforts. The survey also found that a lack of specialized contractors, a lack of digital skills in the workforce, and uncertainty about how to improve energy efficiency are significant barriers to investing in energy efficiency improvements (ABB Ltd, 2022).

The conclusion of a study undertaken by the International Energy Agency suggests that this situation could be improved with digitalization (IEA, 2017). Expert systems (ESs) are one type of advanced computer technology that have emerged from research in the field of artificial intelligence and are designed specifically to assist in decision-making (DeTore, 1989). They pose a chance to overcome the mentioned barriers by combining expert knowledge and analyzing energy data to automatically identify energy efficiency potentials (Ioshchikhes et al., 2023). In general, ESs - also referred to as knowledge-based systems or inference-based programs - can be described as computer programs that leverage expertise to solve problems and provide advice (DeTore, 1989). Unlike conventional applications, these systems emulate human reasoning by representing human knowledge and employ heuristic or approximate methods to solve problems (Jackson, 1998). The intelligent activity that characterizes ESs is the use of knowledge for their processing and not just information. Information exists by itself without a context, whereas knowledge has the added dimension that something is done to process the information. (DeTore, 1989) Energy data from a machine would be information. Reading measurement data, determining which energy efficiency potentials are present, and drawing conclusions on how to exploit existing energy efficiency potentials is the use of knowledge about energy data. This knowledge must be provided by human experts to develop the ES.

There are several literature reviews covering ESs in general and ESs or artificial intelligence to optimize industrial production. Liao (2005) presents an overview of the development of ESs based on a literature review and a classification of articles from 1995 to 2004 into eleven methods. The respective applications of ESs for the associated methods are also given. This publication thus provides a good overview of the various applications and techniques of ESs. However, Liao (2005) does not address the use of ESs to increase energy efficiency and does not analyze any publications in this area. Biel and Glock (2016) examine literature from 1983 to 2015 dealing with decision support models for energy-efficient production planning. For this purpose, procedures for production systems, energy price laws, and energy efficiency criteria are considered. Although this literature analysis focuses on improving energy efficiency in production, it does not include ESs to achieve this objective. Similar to Liao (2005), Wagner (2017) provides a broad overview of ESs and their applications from 1984 to 2016. In comparison to Liao (2005), Wagner (2017) introduces an indicator to estimate and compare the success of ES applications. Among many other applications of ESs, this publication also deals with those in production. Nevertheless, applications for increasing energy efficiency are not discussed. Kumar (2019) focuses his literature review on applications of ESs in production planning regarding the handling of different products, process planning, tool selection, welding, and product development between 1981 and 2016, but without considering energy efficiency.

Unlike the previously cited reviews, this work focuses on ESs to overcome environmental and economic challenges by improving energy efficiency in manufacturing. Other industrial sectors are also considered to ensure that the scope of the study is not overly restricted and to allow possible conclusions to be drawn about manufacturing. This is done through a systematic literature review (SLR) to give a comprehensive overview of the existing research. Starting from a total of 1692 publications, the 54 most relevant publications are extracted and categorized according to system boundary, manufacturing type, application perspective, application purpose, ES type, and industry.

The remainder of this paper is organized as follows: Section 2 describes the methodological approach of the SLR, which is then applied in section 3. This involves all steps, from focusing the topic to identifying research gaps. Finally, section 4 draws a conclusion and presents proposals for future research.

## 2. Review methodology

To identify relevant publications on ESs for improving energy efficiency in manufacturing, a SLR is conducted. A SLR is a formal approach that aims to reduce bias due to an overly restrictive selection of the available literature and to increase the reliability of the literature selected (Tranfield et al., 2003). Moreover, a SLR can clarify the state of research on a topic and highlight gaps and areas requiring further research (Feak & Swales, 2009). The SLR in this paper follows the approach of Reynolds et al. (2003) as well as Glock and Hochrein (2011), which is complemented with elements of Tranfield et al. (2003) and consists of seven consecutive steps shown in Fig. 1. First, the topic is focused and conceptualized. When focusing the topic (I), categories and their definitions are formulated, as consistent classifications help to ensure reliability when conducting content analyses. In the conceptualization (II), research questions for the review itself as well as inclusion and exclusion criteria are formulated. Moreover, the keywords and the search query string are defined. Subsequently, databases suitable for the topic area are selected and searched (III). For this purpose, databases of previous literature analyses covering the same topic area or suggestions of university libraries for databases of individual research fields can be examined. For the literature search, we only consider metadata (title, abstract, keywords). Thereby we avoid publications which contain the search terms only in the bibliography. After merging hits from different databases, the literature is filtered (IV) by applying the defined inclusion and exclusion criteria. In the next step, the full texts of the remaining publications are thoroughly analyzed (V). This involves extracting and summarizing relevant information from the publications. By performing backward and forward snowballing additional relevant publications are identified. Backward snowballing means using the reference list of selected literature to identify new publications to include. Forward snowballing, on the other hand, is carried out to find publications that cite the selected publication. The added publications pass through steps (III) to (V) again. During the synthesis (VI) of the literature essential characteristics are categorized. Finally, the findings are summarized, conclusions are derived with regard to the research questions, and research gaps are identified (VII).

## 3. Literature review

### 3.1. Focusing the topic

To focus the topic, relevant categories and associated dimensions are identified as shown in Fig. 2. The classification scheme is based on Walther and Weigold (2021) for the field of energy modeling in the manufacturing industry. In addition to the focusing of the topic, the classification supports defining keywords for the search query string and the literature analysis.

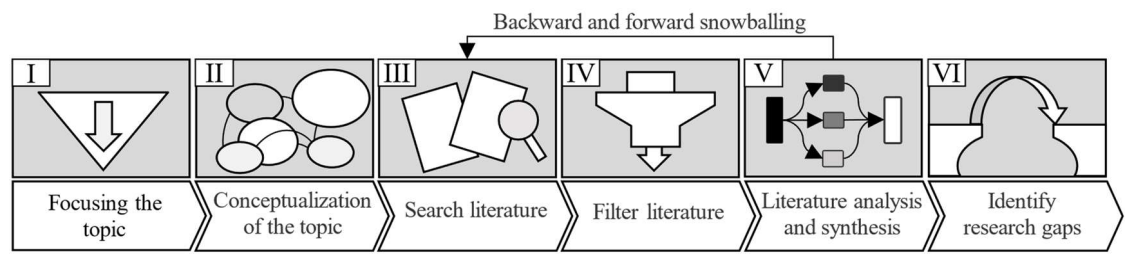

**Fig. 1.** Methodological steps for a systematic literature review (own illustration adapted from Panten (2019) based on Reynolds et al. (2003)).

| System boundary | Manufacturing type | Application perspective | Application focus | Application purpose | Expert system type |
|---|---|---|---|---|---|
| Factory | Job-shop manufacturing | Engineering | Energy efficiency | Transparency | Rule-based expert system |
| Manufacturing cell/ line | Repetitive manufacturing | Process planning | Energy flexibility | Optimization | Fuzzy expert system |
| Machine | Variant manufacturing | Operation | | Prediction | Machine learning based expert system |
| Component | Serial manufacturing | | | Forecasting | Hybrid expert system |
| Process | Mass manufacturing | | | | |

**Fig. 2.** Categories and dimensions for classification.

### 3.1.1. System boundary

In the context of energy-related ES applications in industry, five dimensions are distinguished regarding the system boundary.

- Factory: Distinct physical entity containing multiple devices (Duflou et al., 2012).
- Manufacturing cell/ line: Logical organization of multiple machines to achieve a better division of labor (Westkämper, 2006).
- Machine: Entity required to perform a specific production task (Weck, 2006).
- Component: Individual parts or consumers of a machine which represent the lowest hierarchical level for energy metering (Kara et al., 2011).
- Process: Value-adding and non-value-adding technical operations (Westkämper, 2006).

### 3.1.2. Manufacturing type

Manufacturing systems can be characterized by their manufacturing type, which is determined by the frequency of repetition of a product or product group to be manufactured (Westkämper, 2006).

- Job-shop manufacturing: Custom manufacturing, i.e. according to customer requirements, in which products are only manufactured once.
- Repetitive manufacturing: Products are manufactured at irregular intervals. If orders are repeated, less preparation is required.
- Variant manufacturing: Similar products of the same basic type, which generally involve similar manufacturing effort.
- Serial manufacturing: Mostly contract manufacturing of standardized products in limited quantities.
- Mass manufacturing: Manufacturing large quantities for an anonymous market. High initial investment costs, but cost-effective in relation to the sum of manufactured products.

### 3.1.3. Application perspective

Adapted from Walther and Weigold (2021), the application perspective subdivides the phases in which an ES is useful.

- Engineering: ESs are used for energy-optimized design at a high level of abstraction, e.g. by supporting the selection of sustainable technologies.

- Process planning: The objective of applying ESs in this phase is to plan and optimize manufacturing processes with regard to energy efficiency prior to actual operation, e.g. by optimizing parameters.
- Operation: In the phase in which the actual manufacturing process takes place, ESs are utilized to improve the energy efficiency of the operation, e.g. by detecting inefficient operating points.

*3.1.4. Application focus*

Two dimensions can be distinguished on which studies in the field of energy optimization focus.

- Energy efficiency: Refers to the "relationship between the results achieved and the resources used, where resources are limited to energy" (ISO, 2017).
- Energy flexibility: Describes the "ability of a production system to adapt quickly and in a process-efficient way to changes on the energy market" (Verein Deutscher Ingenieure e.V, 2020).

*3.1.5. Expert system type*

ESs include a user interface, a knowledge base, and an inference engine. The interactions between humans and ESs happen via the user interface, e.g. to define a problem (input) or to read out the results (output). The knowledge base is the programmed knowledge of the expert and contains all pertinent facts and relationships as well as the rules or heuristics to solve problems. The inference engine is the "intelligent" part of an ES which can apply the knowledge from the knowledge base to solve the problem. (DeTore, 1989)

The way in which knowledge is represented within the knowledge base and how it is processed within the inference engine differs depending on the ES type. The typification of ESs in this paper is based on the categories in Liao (2005).

- Rule-based expert system: A rule-based ES represents information in the form of IF-THEN rules. These rules are applied to perform operations on data in order to reach a conclusion (Liao, 2005).
- Fuzzy expert system: Fuzzy ESs are characterized by dealing with uncertainties using fuzzy logic. While rule-based ESs only allow conditions or conclusions that are either true or false, fuzzy ESs allow also conditions or conclusions that are partly true or false. This approach is based on the premise that human experts often decide without precisely quantified information (Liao, 2005).
- Machine-learning-based expert system: This type of ES uses machine leaning (ML) as its "intelligent" component to solve problems. Like ESs, ML belongs to the domain of artificial intelligence and combines a collection of data-driven algorithms that can learn from data without being explicitly programmed. ML also includes deep learning and reinforcement learning. (Ongsulee, 2017)
- Hybrid expert system: Hybrid ESs are a combination of several previously mentioned types or a previously mentioned type with a further approach. Further approaches can be mathematical optimization methods or physical simulation models.

*3.1.6. Application purpose*

There are several reasons for the application of ESs that can contribute to energy efficiency improvements.

- Transparency: To reduce energy consumption in industry, stakeholders need a sufficient level of energy transparency to create a meaningful basis for decision-making (Posselt, 2016).
- Optimization: Optimization in the context of this work means improving energy efficiency as far as necessary and feasible.
- Prediction: Prediction means determining unknown values from known inputs. For energy analysis, this means that the available observations at time $t$ are used to predict the energy or energy efficiency at the same time $t$. (Walther & Weigold, 2021)

- Forecasting: Statements are made about the future. In energy analyses, future values $t + x$ for energy or energy efficiency are estimated based on current and/or past information at time $t$. (Box et al., 2015)

## 3.2. Conceptualization of the topic

In the conceptualization, research questions are raised, inclusion and exclusion criteria are defined, and keywords as well as the search query string are formulated.

### 3.2.1. Research questions
The following research questions (RQ) emerged for the SLR, which are intended to lead to the identification of research gaps:

RQ1  Which industries deploy ESs to increase energy efficiency?
RQ2  How have ESs been applied in the manufacturing industry to enhance energy efficiency?
RQ3  How are ESs for improving energy efficiency in industry structured and implemented?
RQ4  How are ESs for improving energy efficiency in industrial applications developed?

### 3.2.2. Inclusion and exclusion criteria
Inclusion (IC) and exclusion (EC) criteria are defined in advance to ensure consistent filtering within the literature search. Only those publications that pass this selection process are analyzed in depth.

(IC1) Studies written in English or German
(IC2) Reviewed studies
(IC3) Online full text availability
(IC4) Empirical studies with a focus on ESs to improve energy efficiency in industry

(EC1) Studies that do not meet the inclusion criteria
(EC2) Duplicates
(EC3) Excerpts from research results
(EC4) Surveys or reviews (however, should they pertain to this review, they are integrated into section 1 to address related work)

### 3.2.3. Keywords and search query string
The keywords and the search query string are determined based on the classification in Fig. 2. We focus on industrial applications and do not restrict the search to a specific manufacturing type, application perspective, or application purpose. However, to find literature that is more relevant to the objective of this review, the search is limited by specifying the application focus and artificial intelligence technique. For the search query, keywords and their synonyms are concatenated with "OR" operators to broaden the search. By using asterisks (*), words sharing a common root are included in the search as well. The search is restricted by using "AND" operators. The chosen keywords and their logical connections are presented in Table 1.

## 3.3. Search and filter literature

Literature databases were selected based on dos Santos et al. (2023), Panten (2019) as well as Solnørdal and Foss (2018), which are literature reviews related to energy efficiency and flexibility in industry. Thus, the five electronic databases selected are *Web of Science* (https://www.webofscience.com), *IEEE Xplore* (https://ieeexplore.ieee.org), *ScienceDirect* (https://www.sciencedirect.com), *SpringerLink* (https://link.springer.com/) and *WorldCat*

(https://www.worldcat.org). The functionality and syntax of these databases differ slightly. *ScienceDirect*, for example, does not support wildcards such as asterisk (*), other databases including *SpringerLink* do not allow limiting the search to the abstract, so the search string has to be adjusted for a feasible filtering. The search queries used, and the number of results found are listed in Table 1. Fig. 3 provides an overview of the literature search and filtering process. A total of 1692 publications are found by entering the search query. When checking the titles, 1554 publications are filtered out according to the inclusion and exclusion criteria, so that only 138 results remain. This number decreases to 61 articles after examining the abstract. By screening the full texts, another 18 publications are excluded. However, 11 publications are added through the snowball search. This results in a total of 54 publications for a detailed analysis.

Table 1. Databases and applied search queries on August 31, 2023.

| Database | Search query | Number of results |
| --- | --- | --- |
| Web of Science | ((TS=(energ*) OR TS=(load) OR TS=(electri*) OR TS=(power)) AND (TS=(industr*) OR TS=(manufactur*)) AND (TS=(efficien*)) AND (TS=(expert system))) | 807 |
| ScienceDirect | (energy OR load OR electricity OR power) AND (industry OR manufacturing) AND (efficiency) AND (expert system) | 108 |
| IEEE Xplore | ("All Metadata": energ* OR load OR electri* OR power) AND ("All Metadata": industr* OR manufactur*) AND ("All Metadata": efficien*) AND ("All Metadata": expert system) | 121 |
| SpringerLink | ("energy efficiency") AND (industr* OR manufactur*) AND ("expert system") | 371 |
| WorldCat | (energy efficiency) AND (industr* OR manufactur*) AND ("expert system") | 285 |

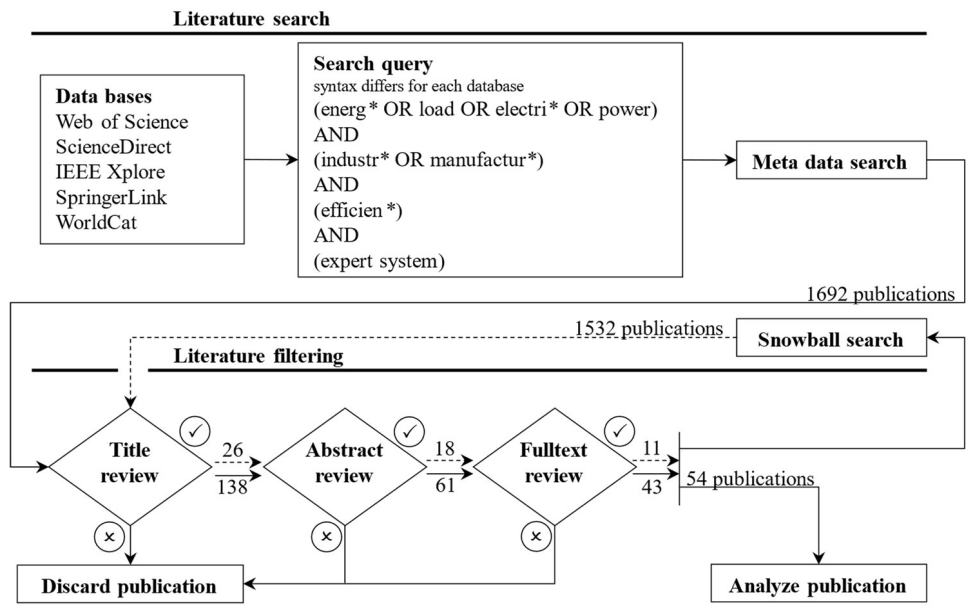

**Fig. 3:** Overview of the literature search and filtering.

*3.4. Literature analysis and synthesis*

Within the scope of the literature analysis and synthesis, initially descriptive information such as the publishing year, the document type, and the ES applications are extracted from the 54 selected publications. Each publication is assigned an identifier (ID) for subsequent analyses. The IDs are characterized by a number and an index, whereby the number is a consecutive sequence starting with the oldest and ending with the most recent publication. The index contains the relevance groups defined by the application areas of the ESs. Group A includes ESs within manufacturing and group B for other industries.

Table 1. Overview of analyzed publications.

| ID | Reference | Document type | Expert system application |
|---|---|---|---|
| $P_A1$ | (Keviczky et al., 1987) | Proceedings paper | Assists operators in setting the technological parameters for cement grinding to optimal working points |
| $P_B2$ | (Miyayama et al., 1991) | Proceedings paper | Supports the control of coal-fired boiler combustion with the aim of high-efficiency operation |
| $P_A3$ | (Zhu et al., 1992) | Journal article | Supports multi-criteria decision-making for analyzing energy systems by simultaneously considering economic and energy-related criteria |
| $P_A4$ | (Tuma et al., 1996) | Journal article | Selects control strategies to influence energy and material flows in interconnected production units and processes |
| $P_A5$ | (Kontopoulos et al., 1997) | Journal article | Guides furnace operators to reduce sensible heat losses by operating with a marginal excess air |
| $P_A6$ | (Dunning et al., 1999) | Journal article | Assists manufacturers in assessing their facilities, minimizing waste, and improving energy efficiency |
| $P_A7$ | (Lau et al., 2008) | Journal article | Forecast energy consumption change to reduce the uncertainty, inconvenience, and inefficiency resulting from variations in the production factors |
| $P_B8$ | (Lin et al., 2008) | Journal article | Enhances the three-phase balance of distribution systems and reduces the neutral current of distribution feeders |
| $P_B9$ | (Soyguder & Alli, 2009) | Journal article | Predicts the fan speed and the damper gap rates of a heating, ventilating, and air-conditioning system |
| $P_B10$ | (Beck & Göhner, 2012) | Journal article | Enables a fully automated, user-centric energy cost analysis for industrial automation systems |
| $P_A11$ | (Garcia, 2012) | Journal article | Diagnoses the fouling condition and heat transfer efficiency of a heat exchanger |
| $P_A12$ | (Klein et al., 2012) | Journal article | Enables identifying energy and operating cost reductions for new and existing filtering installations |
| $P_B13$ | (Roy & Pratihar, 2012) | Journal article | Predicts specific energy consumption and stability margin in case of ascending and descending motions of a six-legged robot |
| $P_A14$ | (Monedero et al., 2012) | Journal article | Increases the energy efficiency of a petrochemical plant by helping the operator take decisions to optimize future operating points |
| $P_B15$ | (Djatkov et al., 2014) | Journal article | Assesses and supports improving the efficiency of agricultural biogas plants based on specific performance figures |
| $P_A16$ | (Do et al., 2014) | Journal article | Supports selecting thermal process technologies in the food industry considering sustainability |
| $P_A17$ | (Mitra et al., 2014) | Proceedings paper | Predicts furnace heating cycles with the objective of achieving the desired product quality with minimal energy utilization |
| $P_B18$ | (Bagheri et al., 2015) | Journal article | Predicts and evaluates the energy efficiency of a forced-convection solar dryer |
| $P_A19$ | (Iqbal et al., 2015) | Journal article | Suggests suitable settings for cutting parameters leading to a trade-off among energy consumption, tool life, and productivity in a machining process |
| $P_B20$ | (Ochoa & Capeluto, 2015) | Journal article | Diagnoses suitable technologies to adapt energy façade systems for residential buildings |
| $P_A21$ | (P. P. Singh & Madan, 2016) | Journal article | Assesses the sustainability performance of a die-casting process plan by determining $CO_2$ emissions, solid waste, and energy use |
| $P_A22$ | (S. Singh et al., 2016) | Proceedings paper | Assesses the sustainable manufacturing performance in small and medium-sized enterprises |
| $P_A23$ | (Teja et al., 2016) | Proceedings paper | Controls and optimizes a triple string rotary cement kiln to improve energy efficiency and increase production while ensuring good quality |
| $P_A24$ | (Cai & Shao, 2017) | Proceedings paper | Determines the initial probability of the energy efficiency state of milling process for further classification |
| $P_A25$ | (Deng et al., 2018) | Journal article | Optimizes cutting parameters for machine tools based on cutting process cases |
| $P_A26$ | (Cheng & Liu, 2018) | Journal article | Optimizes parameters of an injection molding process and measures the energy savings |
| $P_B27$ | (Fernández-Cerero et al., 2018) | Journal article | Applies energy policies based on shutting machines off in order to reduce data-center energy consumption |
| $P_B28$ | (Garofalo et al., 2018) | Journal article | Assess the impact of soil treatment and mineral nitrogen on the energy performance and efficiency of sweet sorghum in the bioethanol supply chain |
| $P_A29$ | (Burow et al., 2019) | Proceedings paper | Optimizes the energy efficiency of a running production process by utilizing scattered data from multiple distributed sources |
| $P_B30$ | (Debnath et al., 2019) | Journal article | Optimized parameters of solar air collectors with corrugated plates under different climatic conditions |
| $P_A31$ | (Simmons et al., 2019) | Proceedings paper | Improves the stability of a cement mill and reduces energy consumption for a cement mill circuit |
| $P_A32$ | (Junfeng Wang et al., 2019) | Journal article | Switches machines in manufacturing systems with serial, disassembly, and assembly workstations into sleep state at an appropriate opportunity |
| $P_A33$ | (Zhao et al., 2019) | Journal article | Schedules engine remanufacturing at multi-machine level to minimize remanufacturing energy consumption |
| $P_A34$ | (Buccieri et al., 2020) | Journal article | Preliminarily diagnoses energy efficiency potential in Brazilian industrial plants and manages knowledge in organizational environments |
| $P_A35$ | (Grigoras & Neagu, 2020) | Journal article | Supports energy consumption management decisions for small and medium-sized enterprises |
| $P_B36$ | (Khayum et al., 2020) | Journal article | Predicts the performance of an anaerobic reactor for the production of biogas using cow dung with spent tea waste in different proportions |
| $P_B37$ | (Kim & Nam, 2020) | Journal article | Presents multiple design decision paths for small- and mid-sized buildings that pursue a balance between economic value and energy performance |
| $P_A38$ | (Petruschke et al., 2021) | Proceedings paper | Supports energy efficiency improvements of machine tools by providing energy efficiency measures and their evaluation |
| $P_A39$ | (Qu & You, 2021) | Journal article | Predicts the situation of sintering furnaces to diagnose faults with the aim of reducing energy consumption during smelting |
| $P_B40$ | (Davydenko et al., 2022) | Journal article | Supports scheduling the charging of an electric vehicle fleet by assigning each electric vehicle a priority for connection to the charging station |
| $P_A41$ | (Ioshchikhes et al., 2022) | Proceedings paper | Identifies system leakage and increased flow resistance in hydraulic systems and quantifies potential energy savings |
| $P_A42$ | (Kalayci et al., 2022) | Journal article | Increases production performance of industrial mixers in terms of product quality, homogeneity, time, and energy savings |
| $P_B43$ | (Li et al., 2022) | Journal article | Predicts existing building commissioning outcomes for various types of public buildings |
| $P_A44$ | (Mendia et al., 2022) | Journal article | Characterizes the nominal performance of a factory in terms of production and energy consumption |
| $P_A45$ | (Rahman et al., 2022) | Journal article | Proposes manufacturing projects an energy-efficient resource-constrained project scheduling plan embedded with a supplier selection strategy |
| $P_B46$ | (Rahmati & Nikbakht, 2022) | Journal article | Predicts the thermohydraulic performance of solar air heater roughened with inclined broken roughness |
| $P_B47$ | (Taleb et al., 2022) | Journal article | Determines the duration of the sleep mode and the sent data rate of a healthcare monitoring system for power consumption optimization |
| $P_A48$ | (X. Wang et al., 2022) | Journal article | Finds solutions to maintain the balance between resource consumption and process indices for a double-stream alumina digestion process |
| $P_A49$ | (Choudhury & Chandrasekaran, 2023) | Journal article | Optimizes parameters during electron beam welding of Inconel 825 for minimizing net input energy without compromising product quality |
| $P_B50$ | (Duman & Seckin, 2023) | Journal article | Reduces the fuel consumption of cabin heaters in emergency shelters and enhances their efficiency |
| $P_A51$ | (Ioshchikhes et al., 2023) | Proceedings paper | Analyzes the energy consumption of chamber cleaning machines and provides measures to improve energy efficiency |
| $P_A52$ | (Jinling Wang et al., 2023) | Journal article | Controls and optimizes grinding process parameters based on monitored power data |
| $P_A53$ | (Perera et al., 2023) | Journal article | Aids metal manufacturing facilities in selecting binder jetting, direct metal laser sintering, or CNC machining |
| $P_B54$ | (Reddy et al., 2023) | Journal article | Optimizes parameters of sand-coated solar air collectors for different climatic conditions |

### 3.4.1. Data sources and publication trend

The 54 publications include a total of 43 journal articles and 11 proceedings papers, whereby both document types are distributed rather evenly throughout the years. As presented in Fig. 4, the publications are not concentrated on a small number of journals or conference proceedings but are spread across a total of 42 sources. The highest concentration can be observed in *Expert Systems and Application* with 6 publications. Other sources listed in Fig. 4 are represented by 2 publications each. The remaining 34 journals and conference proceedings are each represented by a single publication.

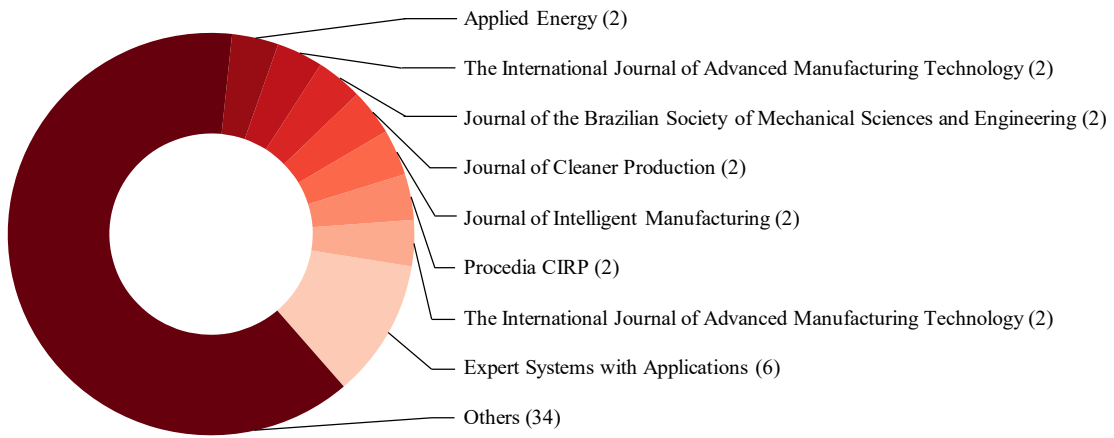

**Fig. 4:** Journal and proceedings distribution.

Fig. 5 shows the number of analyzed publications throughout the years, separated by both relevance groups. No time restriction was set for the literature search, which means that publications from 1987 to 2023 could be reviewed. The year 2023 is not fully covered since the search was carried out in August 2023. Up to 2009, a small number of studies were published. From 2012 onwards, an increase in the average number of publications over the years can be observed for both groups. Furthermore, it is noticeable that the average number of publications in Group A has increased significantly more than in Group B over the last ten years.

### 3.4.2. Authors' country distribution

The authors' affiliation is an additional source of information that is extracted from the metadata of the selected publications. Fig. 6 visualizes the authors' distribution across the individual countries. Most authors are affiliated with China (38). Followed by a double-digit number of authors from India (25), Germany (24), the United States (14) and Spain (14). The continents where most authors are concentrated are Asia (106), Europe (77), and North America (14). The remaining authors originate from South America (3) and Australia (4). There is no observable trend of a particular change in the frequency of publication from individual countries or continents over certain time periods.

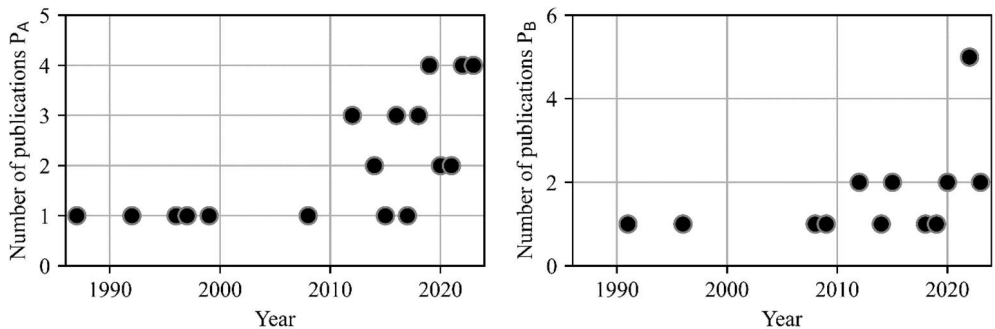

**Fig. 5:** Publishing years of the analyzed publications.

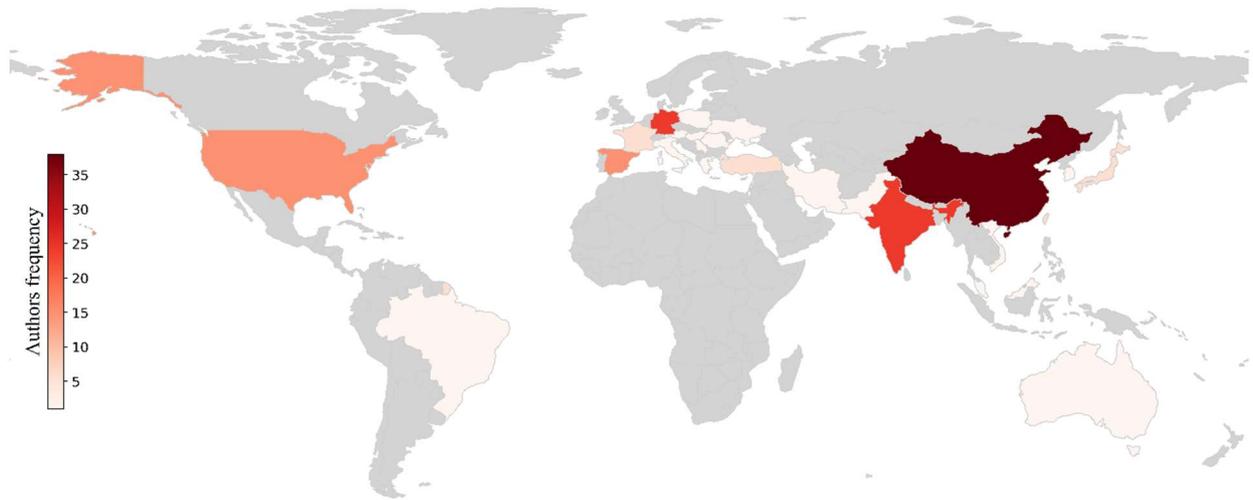

**Fig. 6.** Authors' country distribution.

*3.4.3. Literature categorization*

In this section, publications in relevance group A are categorized in accordance with Fig. 2 by system boundary, manufacturing type, application perspective, application purpose, and ES type. Since publications in relevance group B do not address manufacturing, they are only categorized by application perspective, application purpose, and ES type.

Regarding the system boundary (Table 2), all dimensions are covered, whereby the actual manufacturing process dominates with one third of the publications in group A. In terms of the manufacturing type (Table 3), it can be observed that the use of ESs is concentrated on serial and mass manufacturing, i.e. in larger quantities. Concerning the application perspective (Table 4), most of the studies address process planning and operation. The application purpose of ESs (Table 5) is primarily to increase energy transparency and perform energy optimization, but there are also some publications that predict or forecast information in the context of energy efficiency. Among the ES types (Table 6), pure ML-based ESs are the least common. The other three dimensions, rule-based ES, fuzzy ES, and hybrid ES are approximately evenly represented. Hybrid ESs are combinations of rule-based and fuzzy ESs ($P_A16$), rule-based and ML-based ESs ($P_A11$, $P_A35$, $P_A38$, $P_A52$), fuzzy and ML-based ESs ($P_A4$, $P_B9$, $P_B13$, $P_B18$, $P_B54$), rule-based ESs coupled with simulation models ($P_B10$, $P_A41$) and rule-based or fuzzy ESs using mathematical optimization algorithms ($P_A25$, $P_A45$, $P_A48$, $P_A49$).

Table 2: Categorization of group A publications by system boundaries.

|  | Factory | Manufacturing cell/ line | Machine | Component | Process |
|---|---|---|---|---|---|
| Publication | $P_A6$; $P_A7$; $P_A22$; $P_A34$; $P_A35$; $P_A44$; $P_A45$ | $P_A4$; $P_A16$; $P_A21$; $P_A29$; $P_A31$; $P_A32$; $P_A33$; $P_A53$ | $P_A14$; $P_A17$; $P_A19$; $P_A28$; $P_A38$; $P_A52$ | $P_A3$; $P_A11$; $P_A17$; $P_A38$; $P_A41$ | $P_A1$; $P_A5$; $P_A12$; $P_A17$; $P_A23$; $P_A24$; $P_A25$; $P_A26$; $P_A38$; $P_A39$; $P_A48$; $P_A49$; $P_A51$ |

Table 3: Categorization of group A publications by manufacturing type.

| | Job-shop manufacturing | Repetitive manufacturing | Serial manufacturing | Mass manufacturing |
|---|---|---|---|---|
| Publication | $P_A53$ | $P_A35$; $P_A53$ | $P_A4$; $P_A7$; $P_A16$; $P_A19$; $P_A21$; $P_A24$; $P_A25$; $P_A26$; $P_A32$; $P_A33$; $P_A38$; $P_A39$; $P_A44$; $P_A51$; $P_A52$; $P_A53$ | $P_A1$; $P_A4$; $P_A5$; $P_A7$; $P_A14$; $P_A17$; $P_A19$; $P_A21$; $P_A23$; $P_A24$; $P_A25$; $P_A26$; $P_A28$; $P_A31$; $P_A32$; $P_A33$; $P_A38$; $P_A39$; $P_A44$; $P_A48$; $P_A51$; $P_A52$; $P_A53$ |

Table 4: Categorization of group A publications by application perspective.

| | Engineering | Process planning | Operation |
|---|---|---|---|
| Publication | $P_A3$; $P_A16$; $P_A35$ | $P_A1$; $P_A4$; $P_A6$; $P_A12$; $P_A16$; $P_A19$; $P_A21$; $P_A22$; $P_A25$; $P_A26$; $P_A28$; $P_A29$; $P_A33$; $P_A34$; $P_A45$; $P_A48$; $P_A52$; $P_A53$ | $P_A1$; $P_A5$; $P_A7$; $P_A11$; $P_A14$; $P_A17$; $P_A23$; $P_A24$; $P_A31$; $P_A32$; $P_A38$; $P_A39$; $P_A41$; $P_A44$; $P_A49$; $P_A51$; $P_A52$ |

Table 5: Categorization of publications by application purpose.

| | Transparency | Optimization | Prediction | Forecasting |
|---|---|---|---|---|
| Publication | $P_A1$; $P_A5$; $P_A6$; $P_B10$; $P_A11$; $P_A12$; $P_B13$; $P_B15$; $P_A16$; $P_A21$; $P_A22$; $P_A24$; $P_A26$; $P_B30$; $P_A31$; $P_A34$; $P_A35$; $P_A38$; $P_B40$; $P_A41$; $P_B43$; $P_A44$; $P_B46$; $P_A51$; $P_A52$; $P_A53$; $P_B54$ | $P_A1$; $P_B2$; $P_A3$; $P_A4$; $P_B8$; $P_B9$; $P_B10$; $P_A12$; $P_A14$; $P_B15$; $P_A16$; $P_A17$; $P_A19$; $P_B20$; $P_A23$; $P_A25$; $P_A26$; $P_B27$; $P_A28$; $P_A29$; $P_B30$; $P_A31$; $P_A32$; $P_A33$; $P_A35$; $P_B37$; $P_A38$; $P_A39$; $P_B40$; $P_A41$; $P_B42$; $P_B43$; $P_A45$; $P_B47$; $P_A48$; $P_A49$; $P_B50$; $P_A51$; $P_A52$; $P_B54$ | $P_A12$; $P_B13$; $P_B18$; $P_A19$; $P_B27$; $P_A28$; $P_B36$; $P_A41$; $P_B43$ | $P_A6$, $P_A7$; $P_A11$; $P_A35$; $P_A51$ |

Table 6: Categorization of publications by expert system type.

| | Rule-based ES | Fuzzy ES | ML-based ES | Hybrid ES |
|---|---|---|---|---|
| Publication | $P_A1$; $P_A3$; $P_A5$; $P_A6$; $P_B8$; $P_A12$; $P_A17$; $P_B20$; $P_A21$; $P_A24$; $P_A26$; $P_B27$; $P_A29$; $P_A31$; $P_A34$; $P_B37$; $P_B43$; $P_A53$ | $P_B2$; $P_A7$; $P_B15$; $P_A19$; $P_A22$; $P_A23$; $P_A28$; $P_B30$; $P_A32$; $P_A33$; $P_B36$; $P_A39$; $P_B40$; $P_B42$; $P_B46$; $P_B47$; $P_B50$; $P_A51$ | $P_A14$; $P_A44$ | $P_A4$; $P_B9$; $P_B10$; $P_A11$; $P_B13$; $P_A16$; $P_B18$; $P_A25$; $P_A35$; $P_A38$; $P_A41$; $P_A45$; $P_A48$; $P_A49$; $P_A52$; $P_B54$ |

*3.5. Identify research gaps*

Answers to the research questions raised in section 3.2.1 are provided in the following and discussed to identify research gaps.

### 3.5.1. Classification of ESs by industries

For answering RQ1, in which industries ESs are used to increase energy efficiency, the classification of economic sectors of the Federal Statistical Office in Germany is considered. The classification is hierarchically structured and is divided into sections, which are separated into divisions (Statistisches Bundesamt, 2023).

Table 7 summarizes the analyzed publications according to this classification scheme. We can classify the publications into six industries. Some publications cannot be assigned clearly because they deal with cross-sectional technologies or do not provide any information about the exact industry. There is a noticeable concentration of the use of ESs in the manufacturing of metal goods. Furthermore, Table 7 shows again that with 65 % the majority of ESs for improving energy efficiency are deployed in manufacturing (relevance group A).

Table 7: Industry classification.

| Industry | Division | Publication |
|---|---|---|
| Manufacturing | Food and feed manufacturing | $P_A12$; $P_A16$ |
| | Textile manufacturing | $P_A4$; $P_A7$ |
| | Chemical product manufacturing | $P_A14$; $P_A28$ |
| | Plastic goods manufacturing | $P_A26$ |
| | Cement manufacturing | $P_A1$; $P_A12$; $P_A23$; $P_A31$ |
| | Iron and steel manufacturing and processing | $P_A17$; $P_A21$; $P_A49$ |
| | Aluminum manufacturing and initial processing | $P_A5$; $P_A48$ |
| | Metal goods manufacturing | $P_A19$; $P_A24$; $P_A25$; $P_A38$; $P_A39$; $P_A41$; $P_A51$; $P_A52$; $P_A43$ |
| | Automotive and automotive parts manufacturing | $P_A33$; $P_A35$; $P_A44$; |
| | Others | $P_A3$; $P_A6$; $P_A11$; $P_A22$; $P_A29$; $P_A32$; $P_A34$; $P_A45$ |
| Energy supply | Electric power distribution | $P_B8$ |
| | Gas production | $P_B15$; $P_B36$ |
| | Heating and cooling supply | $P_B2$ |
| Provision of professional, scientific, and technical services | Architectural and engineering offices | $P_B20$; $P_B37$; $P_B43$ |
| Information Services | Data processing, hosting, and related activities | $P_B27$ |
| Transport and storage | Passenger transport by land | $P_B40$ |
| Healthcare | Hospitals | $P_B47$ |
| Others | | $P_B9$; $P_B10$; $P_B13$; $P_B18$; $P_B30$; $P_B42$; $P_B46$; $P_B50$; $P_B54$ |

### 3.5.2. Utilizations of ESs in industry

Regarding the question of how ESs are applied to enhance energy efficiency in manufacturing (RQ2), it can be observed that ESs with a wider system boundary are more transferable than others with a more restricted system boundary. Thus, Dunning et al. (1999), Lau et al. (2008), S. Singh et al. (2016), Buccieri et al. (2020) as well as Grigoras and Neagu (2020) describe ESs with system boundaries covering entire factories but are rather utilized for preliminary or generic energy analyses. Some other publications with a more limited system boundary such as Garcia (2012), Do et al. (2014), P. P. Singh and Madan (2016), Petruschke et al. (2021) and Ioshchikhes et al. (2023) are applicable to similar use cases with few or no adjustments. The remaining publications in relevance group A describe ESs that are largely limited to individual case studies and cannot be transferred to other machines, systems or processes without further modification.

The classification in Table 4 also shows that the utilization of ESs is strongly focused on either planning manufacturing processes regarding energy efficiency before actual operation or optimization during operation. ESs are deployed less frequently for energy-efficient planning of new energy systems. Furthermore, according to Table 5, the primary purpose of ESs is to create energy transparency as a basis for subsequent improvement processes as well as to contribute directly to improving energy efficiency.

*3.5.3. Structure and implementations of ESs*

RQ3 addresses the structure and implementation of ESs. All ESs described within the analyzed 54 publications feature a knowledge base, an inference engine, and a user interface, which are interconnected as shown in Fig. 7. Some of the publications, for instance, Buccieri et al. (2020) as well as Grigoras and Neagu (2020), additionally include distinct knowledge acquisition and explanation modules.

The expertise stored in the knowledge base can be represented in various forms. These include mathematical formulas (Keviczky et al., 1987), historical measurement data (Miyayama et al., 1991) physical and technical relationships (Tuma et al., 1996), rules (Lin et al., 2008), and facts (Do et al., 2014). Furthermore, as outlined by Buccieri et al. (2020), the knowledge base can be divided into short-term and long-term memory. Short-term memory stores information for particular use cases for which ESs are applied. Whereas the long-term memory stores more general relationships, which correspond to the heuristic knowledge of human experts. Knowledge can be acquired using different techniques. Literature surveys are explicitly mentioned by Do et al. (2014) and Perera et al. (2023) as well as expert consultations by Do et al. (2014), Buccieri et al. (2020), and Perera et al. (2023). The optional knowledge acquisition module enables the knowledge base to be enriched with new content while the ES is already deployed. Depending on the ES type (Table 6), it varies how the knowledge is represented in the knowledge base and how the inference engine utilizes the knowledge stored in the knowledge base to gain insights or solve problems.

The communication between humans and the ES (human-machine interface) can be realized by different means, including interactive questionnaires (Keviczky et al., 1987), dialog boxes (Dunning et al., 1999), measurement data inputs (Ioshchikhes et al., 2023), text outputs and graphics (Ochoa & Capeluto, 2015). The optional explanation module clarifies the reasoning process in order to make it comprehensible and thus promote acceptance of the findings presented by the ES (Buccieri et al., 2020).

As the analyzed publications reveal, the software implementation of ES can be highly diverse. Keviczky et al. (1987) use Fortran and Assembler as programming languages. More recent publications implement their ESs in Excel (Do et al., 2014), C language (Iqbal et al., 2015), CLIPS (Buccieri et al., 2020), Python (Petruschke et al., 2021), and LabVIEW (Jinling Wang et al., 2023). Deng et al. (2018), Ioshchikhes et al. (2022), and Li et al. (2022) use combinations of different programming languages to integrate simulation models or databases as part of the ESs.

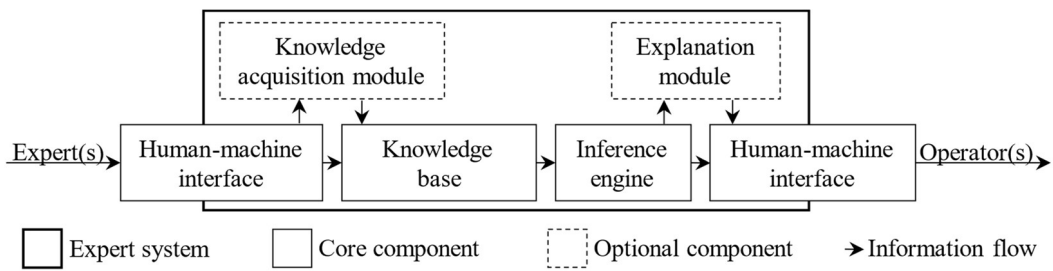

**Fig. 7.** Simplified structure of ESs.

*3.5.4. Development of ESs for industrial applications*

Many analyzed publications follow similar sequential and sometimes iterative steps in the development of their ES. These steps are:

- Situation analysis: Create an overview of the situation and examine the problem to be solved (Lau et al., 2008).
- Parameter identification: Identification of important and interactive parameters and factors, i.e. those that are both controllable (Monedero et al., 2012) and influential (Li et al., 2022).
- Data generation and acquisition: Data is generated and collected using simulation models (Keviczky et al., 1987) or experiments (Iqbal et al., 2015). Experimental data generation is more common in the analyzed publications.
- Data cleaning and preprocessing: Data filtering (Monedero et al., 2012), scaling (Debnath et al., 2019) and further preprocessing such as discriminant analysis (Monedero et al., 2012) or feature extraction (Jinling Wang et al., 2023) is performed.
- Creation of a knowledge base and knowledge representation: Knowledge is gathered and represented as described in section 3.5.3.
- Modeling: Build mathematical models (Li et al., 2022), fuzzy models, or data-driven models such as neural networks (Tuma et al., 1996). The development steps of the modeling process depend on the respective model type.
- Application and validation: The ES can be qualitatively validated via user interviews and quantitatively through case studies (Li et al., 2022).

Buccieri et al. (2020) and Li et al. (2022) specify corresponding development methods. Buccieri et al. (2020) follow an incremental development model exclusively for the computational implementation, without providing a detailed description of the individual phases. Li et al. (2022) base the entire development of their ES on a design science method. The method described by Li et al. (2022) comprises three phases: conceptual design phase, tool development phase, as well as application and validation phase. During the conceptual design, influential factors are selected, calculation rules are defined, and iteratively refined. Subsequently, the computational implementation follows during the tool development. Finally, the ES is validated both qualitatively and quantitatively.

To summarize, various development approaches for ESs can be found with some similarities in the context of improving industrial energy efficiency. Most development approaches, except for those described by Buccieri et al. (2020) and Li et al. (2022), cannot be associated with a superior methodological approach. In the area of manufacturing, no established process model with a detailed description of the individual steps for developing ESs to increase energy efficiency could be found.

## 4. Conclusion

This paper presents a SLR that identifies, summarizes, and analyzes ESs to improve energy efficiency in industry with a focus on manufacturing. The research follows a formal approach consisting of seven consecutive steps, aiming to reduce bias and increase the reliability of the literature selected. This led to 54 publications that were analyzed in depth. To organize the findings, the proposed ESs were classified according to system boundary, manufacturing type, application perspective, application purpose, and ES type. Subsequently, the raised research questions were answered, and research gaps were identified.

Regarding existing literature concerning ESs, the main contribution of this work lies in the special focus on improving energy efficiency in manufacturing. Other industries were also considered to allow a full assessment of manufacturing. We classified this topic into different classes, dimensions, and industries. Furthermore, we examined the structure, implementation, utilization, and development of ESs in this context.

Within the scope of this study, there are some limitations in terms of covering the state of the art in the investigated research area. The exclusive reliance on digital sources and the deliberate overlook of older publications potentially causes a bias toward recent findings. Another limitation results from the language restriction, although only one of the 54 publications analyzed was not written in English.

Nonetheless, this study provides a comprehensive review of the topic area covered. The analysis of the selected publications led to several observations that can be highlighted and discussed. While Fig. 5 illustrates a rise in the number of publications, it offers no insight into whether the observed trend results from an increased interest in the topic or a consequence of the broader expansion of research and scientific output. However, the comparison between the two relevant groups shows a greater growth in the number of publications in group A over the last ten years. This might be a consequence of the increasing attention being paid to sustainability in manufacturing. The authors' country affiliation in Fig. 6 can also be interpreted against the background of various factors. Countries with large populations, a strongly established secondary economic sector, and high spending on research have yielded more publications in the researched area than others. Moreover, the answer given to RQ1 (Table 7) combined with the categorization by manufacturing type (Table 3) suggests that ESs for improving energy efficiency are of particular interest to companies with high energy consumption and large output quantities. This could be related to the initial development effort arising from the lack of established process models (RQ4) and the low transferability of ESs to other use cases (RQ2).

The answers provided to the research questions revealed research gaps that can be addressed in future research. Thus, the question arises as to why ESs are used more frequently in certain industries (RQ1) than in others and what reasons, such as energy consumption, output quantities, or digitalization, play a decisive role in this. Furthermore, it became apparent that in most cases ESs are not transferable to other machines, systems, or processes (RQ2). While advantageous for increasing energy efficiency and preserving knowledge, it also means a high initial development effort for each individual use case, which can pose a barrier to implementation. Exploring the transferability aspect in the development of ESs could be a subject of future research. Regarding the structure and implementation of ESs (RQ3), it has been revealed that although all described ESs have a knowledge base, an inference engine, and a user interface, their implementation is highly diverse. Accordingly, it was also pointed out that there is still no established method for the development of ESs to increase energy efficiency in manufacturing.

**CReditT authorship contribution statement**

**Borys Ioshchikhes:** Conceptualization, Methodology, Investigation, Writing - Original Draft, Visualization. **Michael Frank:** Writing - Review & Editing. **Matthias Weigold:** Supervision.

**Declaration of competing interests**

The authors declare that they have no known competing financial interests or personal relationships that could have appeared to influence the work reported in this paper.

**Data availability**

No data was used for the research described in the article.

**Acknowledgements**

The authors gratefully acknowledge financial support from the German Federal Ministry of Economic Affairs and Climate Action (BMWK) for the project *KI4ETA* (grant agreement No. 03EN2053A) and project supervision by the project management Jülich (PtJ).

20